%% file: root.tex
\documentclass{article}

\usepackage[final]{corl_2020} %

\usepackage{mathptmx} %
\usepackage{amsmath} %
\usepackage{amssymb}  %
\usepackage{verbatim}

\usepackage{booktabs} 
\usepackage{graphicx}
\usepackage{color}
\usepackage{array} %
\usepackage{multirow}
\usepackage{hyperref}
\usepackage{subcaption}
\usepackage{algorithm,algpseudocode}
\usepackage{xr}

\usepackage{paralist}
\usepackage{siunitx}
\usepackage{mathtools}
\usepackage{wrapfig}
\usepackage{newtxtext,newtxmath}

\newcommand{\thisWork}{IV-SLAM}
\newcommand{\mypara}[1]{{\smallskip\noindent \bf #1.}\hspace{0.1in}}

\DeclareMathOperator*{\argmin}{arg\,min}
\DeclareMathOperator*{\argmax}{arg\,max}

\newcolumntype{M}[1]{>{\centering\arraybackslash}m{#1}}

\algnewcommand\algorithmicforeach{\textbf{for each}}
\algdef{S}[FOR]{ForEach}[1]{\algorithmicforeach\ #1\ \algorithmicdo}

\newcommand*{\obs}{\mathbf{z}}
\newcommand*{\reprojErr}{\mathbf{\epsilon}}
\newcommand*{\obsErr}{\tilde{\mathbf{\epsilon}}}
\newcommand*{\errDist}{\tilde{\mathcal{\phi}}}
\newcommand*{\costmap}{\mathbf{I_c}_t}
\newcommand*{\costmapEst}{\hat{\mathbf{I}}_{\mathbf{c_{t}}}}
\newcommand*{\lossDelta}{\delta (\obs_{t, k})}
\newcommand*{\camPose}{\mathbf{T}_{t}^w}
\newcommand*{\camPoseEst}{\hat{\mathbf{T}}_{t}^w}
\newcommand*{\camPoseInv}{\mathbf{T}_{w}^t}
\newcommand*{\camPoseAll}{\mathbf{T}_{1:t}^w}
\newcommand*{\stateVector}{\camPoseAll, \mathbf{M}}

\newcommand*{\stateVecSolution}{{\overset{*}{\mathbf{T}_{1:t}^w}}, \overset{*}{\mathbf{M}}}

\title{IV-SLAM: Introspective Vision for \\Simultaneous Localization and Mapping}

\author{
  Sadegh Rabiee\\
  Department of Computer Science\\
  University of Texas at Austin \\
  United States\\
  \texttt{srabiee@cs.utexas.edu} \\
  \And
  Joydeep Biswas\\
  Department of Computer Science\\
  University of Texas at Austin \\
  United States\\
  \texttt{joydeepb@cs.utexas.edu} \\
}

\makeatletter
\g@addto@macro\normalsize{%
  \setlength\abovedisplayskip{3pt}
  \setlength\belowdisplayskip{3pt}
  \setlength\abovedisplayshortskip{-2pt}
  \setlength\belowdisplayshortskip{1pt}
}
\makeatother

\begin{document}
\maketitle

\begin{abstract}
\input{abstract}

\end{abstract}


\input{introduction}

\input{related_work}

\input{background}

\input{method}

\input{experimental_results}

\input{conclusion}

\clearpage
\acknowledgments{This work was partially supported by the Defense Advanced Research Projects Agency (DARPA) under Contract No. HR001120C0031, and by an unrestricted gift from Amazon. In addition, we acknowledge support from Northrop Grumman Mission Systems\textquotesingle~ University Research Program.}

\bibliography{bibliography}  %

\end{document}

%% file: abstract.tex
Existing solutions to visual simultaneous localization and mapping (V-SLAM)
assume that errors in feature extraction and matching are independent and
identically distributed (i.i.d), but this assumption is known to not be true -- features
extracted from low-contrast regions of images exhibit wider error distributions
than features from sharp corners. Furthermore, V-SLAM algorithms are prone to catastrophic tracking
failures when sensed images include challenging conditions such as
specular reflections, lens flare, or shadows of dynamic objects. To address such
failures, previous work has focused on building more robust visual
frontends, to filter out challenging features. In this
paper, we present  \emph{introspective vision} for SLAM (\thisWork), a
fundamentally different approach for addressing these challenges. \thisWork{}
explicitly models the noise process of reprojection errors from visual
features to be context-dependent, and hence non-i.i.d. We introduce an
autonomously supervised approach for \thisWork{} to collect training data to
learn such a context-aware noise model. Using this learned noise model,
\thisWork{} guides feature extraction to select more features from parts of the
image that are likely to result in lower noise, and further incorporate the
learned noise model into the joint maximum likelihood estimation, thus making it
robust to the aforementioned types of errors. We present empirical results to
demonstrate that \thisWork{} 1) is able to accurately predict sources of error
in input images, 2) reduces tracking error compared to V-SLAM, and 3) increases
the mean distance between tracking failures by more than $70\%$ on challenging
real robot data compared to V-SLAM.

%% file: introduction.tex
\newcommand{\todo}[1]{{\color{red}[#1]}}

Visual simultaneous localization and mapping (V-SLAM) extracts features from
observed images, and identifies correspondences between features across time-steps. By jointly optimizing the re-projection error of such
features along with motion information derived from odometry or inertial
measurement units (IMUs), V-SLAM reconstructs the trajectory of a robot along
with a sparse 3D map of the locations of the features in the world. To accurately
track the location of the robot and build a map of the world, V-SLAM requires
selecting features from static objects, and correctly and consistently
identifying correspondences between features. Unfortunately, despite extensive
work on filtering out bad features~\citep{alt2010rapid, wang2006good,
carlone2018attention} or rejecting unlikely correspondence
matches~\citep{zamir2016generic, ono2018lf, tian2019sosnet}, V-SLAM solutions
still suffer from errors stemming from incorrect feature matches and features
extracted from moving objects. Furthermore, V-SLAM solutions assume that
re-projection errors are independent and identically distributed (i.i.d), an
assumption that we know to be false: features
extracted from low-contrast regions or from regions with repetitive textures
exhibit wider error distributions than features from regions with sharp, locally
unique corners. As a consequence of such assumptions, and the reliance on robust frontends to filter out bad features, even state of the art
V-SLAM solutions suffer from catastrophic failures when encountering challenging
scenarios such as specular reflections, lens flare, and shadows of moving
objects encountered by robots in the real world.

We present \emph{introspective vision} for SLAM (\thisWork), a
fundamentally different approach for addressing these challenges -- instead of
relying on a robust frontend to filter out bad features,
\thisWork{} builds a context-aware \emph{total noise
model}~\cite{triggs1999bundle} that explicitly accounts for heteroscedastic
noise, and learns to account for bad correspondences, moving objects, non-rigid
objects and other causes of errors.
\thisWork{} is capable of learning to identify causes of V-SLAM failures
in an autonomously supervised manner, and is subsequently able to leverage the
learned model to improve the robustness and accuracy of tracking during actual deployments.
Our experimental results 
demonstrate that \thisWork{} 1) is able to accurately predict sources of error
in input images as identified by ground truth in simulation, 2) reduces tracking
error on both simulation and real-world data, and 3)
significantly increases the mean distance between tracking failures when
operating under challenging real-world conditions that frequently lead to
catastrophic tracking failures of V-SLAM.

%% file: related_work.tex
\section{Related Work} \label{related_work}
There exists a plethora of research on designing and learning distinctive image feature descriptors. This includes the classical hand-crafted descriptors such as ORB~\cite{rublee2011orb} and SIFT~\cite{lowe2004distinctive}, as well as the more recent learned ones~\citep{balntas2016learning, mishchuk2017working, tian2017l2, ono2018lf, tian2019sosnet} that rely on Siamese and triplet network architectures to generate a feature vector given an input image patch.
Selecting interest points on the images for extracting these descriptors is traditionally done by convolving the image with specific filters~\citep{bay2006surf} or using first-order approximations such as the Harris detector. More recently, CNN approaches have become popular~\citep{lenc2016learning, savinov2017quad}. \citet{cieslewski2019sips} train a network that given an input image outputs a score map for selecting interest points.
Once features are extracted at selected keypoints, pruning out a subset of them that are predicted to be unreliable is done in different ways.
\citet{alt2010rapid} train a classifier to predict good features at the descriptor level, and \citet{wang2006good} use hand-crafted heuristics to determine good SIFT features.
\citet{carlone2018attention} propose a feature pruning method for visual-inertial SLAM that uses the estimated velocity of the camera to reject image features that are predicted to exit the scene in the immediate future frames.
A line of work leverages scene semantic information in feature selection. \citet{kaneko2018mask} run input images through a semantic segmentation network and limit feature extraction to semantic classes that are assumed to be more reliable such as static objects. \citet{ganti2019sivo} follow a similar approach while taking into account the uncertainty of the segmentation network in their feature selection process. While these approaches benefit from using the contextual information in the image, they are limited to hand-enumerated lists of sources of error. Moreover, not all potential sources of failure can be categorized in one of the semantic classes, for which well-trained segmentation networks exist. Repetitive textures, shadows, and reflections are examples of such sources.
Pruning bad image correspondences once features are matched across frames is also another active area of research.
Forward predicting the motion of the robot by means of accurate learned motion models~\cite{rabiee2019friction} reduces outliers to some extent by limiting the region in the image, where the corresponding match for each feature can lie. The remaining wrong correspondences have traditionally been addressed with RANSAC~\citep{fischler1981random}, and more recently deep learning approaches have been developed~\citep{moo2018learning, sun2020acne}, which use permutation-equivariant network architectures and predict outlier correspondences by processing coordinates of the pairs of matched features. While the goal of these methods is to discard outlier correspondences, not all bad matches are outliers. There is also a grey area of features that for reasons such as specularity, shadows, motion blur, etc., are not located as accurately as other features without clearly being outliers.

Our work is agnostic of the feature descriptor type and the feature matching method at hand. It is similar to the work by \citet{cieslewski2019sips} in that it learns to predict unreliable regions for feature extraction in a self-supervised manner.
However, it goes beyond being a learned keypoint detector and applies to the full V-SLAM solution by exploiting the predicted feature reliability scores to generate a context-aware loss function for the bundle adjustment problem. Unlike available methods for learning good feature correspondences, which require accurate ground truth poses of the camera for training~\citep{moo2018learning, sun2020acne, zamir2016generic}, our work only requires rough estimates of the reference pose of the camera.
\thisWork{} is inspired by early works on introspective vision~\citep{8968176, daftry2016introspective} and applies the idea to V-SLAM.

%% file: background.tex
\section{Visual SLAM} \label{sec:background}
In visual SLAM, the pose of the camera $\camPose \in SE(3)$ is estimated and a 3D map $\mathbf{M} = \left\lbrace \mathbf{p}_k^w | \mathbf{p}_k^w \in \mathbb{R}^{3} , k \in [1, N]   \right\rbrace$ of the environment is built by finding correspondences in the image space across frames.  For each new input image $\mathbf{I}_t$ (or stereo image pair
$(\mathbf{I}_{t,l}, \mathbf{I}_{t,r})$, image features are extracted and matched with those from previous frames. Then, the solution to SLAM is given by
\begin{equation}
\stateVecSolution = \argmax_{\stateVector} P(\stateVector|\mathbf{Z}_{1:t}, \mathbf{u}_{1:t}) = \argmax_{\stateVector} P(\mathbf{Z}_{1:t}|\stateVector) P \left(\camPoseAll | \mathbf{u}_{1:t} \right) \, ,
\label{eq:slam_form}
\end{equation}
where $\mathbf{Z_{1:t}}$ represents observations made by the robot and $\mathbf{u}_{1:t}$ are the control commands and/or odometry and IMU measurements. $P(\mathbf{Z}|\stateVector)$ is the observation likelihood for image feature correspondences, given the estimated poses of the camera and the map $\mathbf{M}$. 
For each time-step $t$, the V-SLAM frontend processes image $\mathbf{I}_t$ to extract features $\obs_{t,k} \in \mathbb{P}^{2}$ associated with 3D map points $\mathbf{p}_k^w$.
The observation error here is the reprojection error of $\mathbf{p}_k^w$ onto the image $\mathbf{I}_t$ and is defined as
\begin{equation}
\reprojErr_{t,k} = \obs_{t,k} - \mathbf{\hat{\obs}}_{t, k}, \quad \mathbf{\hat{\obs}}_{t, k} = \mathbf{A} \left[ \mathbf{R}^t_w | \mathbf{t}^t_w \right] \mathbf{p}_k^w \, ,
\label{eq:reprojection_error}
\end{equation}
where $\mathbf{\hat{\obs}}_{t, k}$ is the prediction of the observation $\obs_{t, k}$, $\mathbf{A}$ is the camera matrix, and $\mathbf{R}^t_w \in SO(3)$ and $\mathbf{t}^t_w \in \mathbb{R}^3$ are the rotation and translation parts of $\camPoseInv$, respectively. In the absence of control commands and IMU/odometry measurements, Eq.~\ref{eq:slam_form} reduces to the bundle adjustment (BA) problem, which is formulated as a nonlinear optimization:
\vspace{-0.5em}
\begin{equation}
\stateVecSolution = \argmin_{\stateVector}
\sum_{t, k} L\left(\reprojErr_{t,k}^T \Sigma_{t,k}^{-1} \reprojErr_{t,k}\right) \, ,
\label{eq:bundle_adjustment}
\end{equation}
where $\Sigma_{t,k}$ is the covariance matrix associated to the scale at which a feature has been extracted and $L$ is the loss function for $P(\mathbf{Z}|\stateVector)$.
The choice of noise model for the observation error $\reprojErr$ has a significant effect on the accuracy of the maximum likelihood (ML) solution to SLAM~\citep{triggs1999bundle}. There exists a body of work on developing robust loss functions~\citep{black1996unification, barron2019general} %
that targets improving the performance of vision tasks in the presence of outliers.
The reprojection error is known to have a non-Gaussian distribution $\mathcal{\phi}$ in the context of V-SLAM due to the frequent outliers that happen in image correspondences~\citep{triggs1999bundle}. It is assumed to be drawn from long-tailed distributions such as piecewise densities with a central peaked inlier region and a wide outlier region. While $\mathcal{\phi}$ is usually modeled to be i.i.d., there exist obvious reasons as to why this is not a realistic assumption.
Image features extracted from objects with high-frequency surface textures can be located less accurately; whether the underlying object is static or dynamic affects the observation error; the presence of multiple similar-looking instances of an object in the scene can lead to correspondence errors. 
In the next section, we explain how \thisWork{} leverages the contextual information available in the input images to learn an improved $\mathcal{\phi}$ that better represents the non i.i.d nature of the observation error.

%% file: method.tex
\section{Introspective Vision for SLAM} \label{method}

\begin{figure*}[t]
  \centering 
  \includegraphics[width=\linewidth ,trim=0 0 0 0,clip]{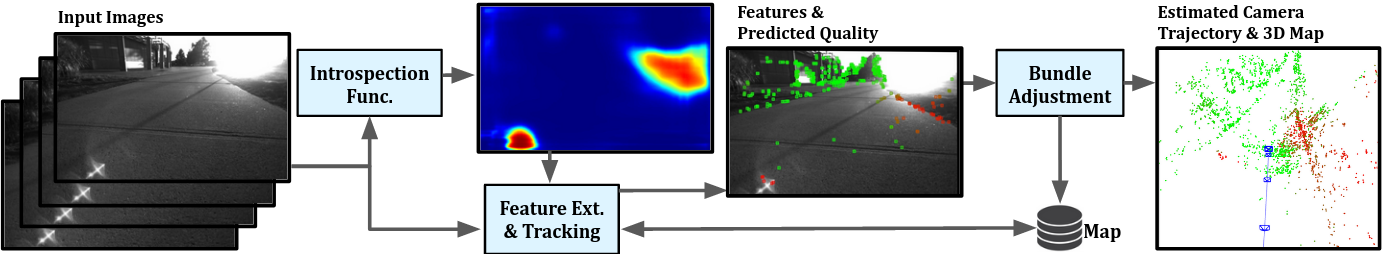}
  \caption{IV-SLAM pipeline during inference.}
  \label{fig:ivslam_pipeline}
  \vspace{-3mm}
\end{figure*}

\thisWork{} builds on top of a feature-based V-SLAM algorithm. It is agnostic of the type of image features or the frontend, hence applicable to a wide range of existing V-SLAM algorithms. \thisWork{}
\begin{inparaenum}[1)]
\item equips V-SLAM with an \emph{introspection function} that learns to predict the reliability of image features, 
\item modifies the feature extraction step to reduce the number of features extracted from unreliable regions of the input image and extract more features from reliable regions, and
\item modifies the BA optimization to take into account the predicted reliability of extracted features.
\end{inparaenum}

\thisWork{} models the observation error distribution to be dependent on the observations, i.e.
$\obs = \mathbf{\hat{\obs}}(\stateVector) + \mathbf{\obsErr(\obs)}$, where $\tilde{\epsilon} \sim \tilde{\mathcal{\phi}}$ and $\tilde{\mathcal{\phi}}$ is a heteroscedastic noise distribution. Let $p_{\errDist}$ be the probability density function (PDF) of $\errDist$, we want
$p_{\errDist} \propto \exp(-L)$, where $L \in \mathcal{L}$ is a loss function from the space of robust loss functions~\citep{barron2019general}. In this paper, we choose $L \in \mathcal{H} \subset \mathcal{L}$, where $\mathcal{H}$ is the space of Huber loss functions and specifically
\vspace{-0.5em}
\begin{equation}
 L_{\delta(\mathbf{z})}(x) =
 \begin{cases}
             x & \text{if $ x < \delta(\mathbf{z})$} \\
             2\delta(\mathbf{z})(\sqrt{x} - \delta(\mathbf{z})/2) & \text{otherwise}
 \end{cases} 
\label{eq:huber_loss}
\end{equation}
where $x \in [0, \infty)$ is the squared error value and $\delta(\mathbf{z})$ is an observation-dependent parameter of the loss function and is correlated with the reliability of the observations.
\thisWork{} uses the introspection function to learn an empirical estimate of $\delta(\mathbf{z})$ such that the corresponding error distribution $\errDist$ better models the observed error values. During the training phase, input images and estimated observation error values are used to learn to predict the reliability of image features at any point on an image. During the inference phase, a context-aware $\delta(\mathbf{z})$ is estimated for each observation using the predicted reliability score, where a smaller value of $\delta(\mathbf{z})$ corresponds to an unreliable observation. The resultant loss function $L_{\delta(\mathbf{z})}$ is then used in Eq.~\ref{eq:bundle_adjustment} to solve for $\camPoseAll$ and $\mathbf{M}$. Fig.~\ref{fig:ivslam_pipeline} illustrates the \thisWork{} pipeline during inference. In the rest of this section, we explain the training and inference phases of \thisWork{} in detail.

\subsection{Self-Supervised Training of ~\thisWork{}}
One of the main properties of \thisWork{} as an introspective perception algorithm is its capability to generate labeled data required for its training in a self-supervised and automated manner. This empowers the system to follow a continual learning approach and exploit logged data during robot deployments for improving perception. In the following, we define the
introspection function in \thisWork{} and explain the automated procedure for its training data collection.

\vspace{-0.5em}
\subsubsection{Introspection Function} \label{sec:introspection_func}
\vspace{-0.5em}
In order to apply a per-observation loss function $L_{\delta(\mathbf{z})}$, \thisWork{} learns an \emph{introspection function} 
$\mathbb{I}:\mathbf{I} \times \mathbb{R}^2 \to \mathbb{R}$ that given an input image $\mathbf{I}_t$ and a location $(i, j) \in \mathbb{R}^2$ on the image, predicts a \emph{cost value} ${c_t}_{i,j} \in [0, 1]$ that represents a measure of reliability for image features extracted at $\mathbf{I}_t(i, j)$. Higher values of ${c_t}_{i,j}$ indicate a less reliable feature.
Our implementation of $\mathbb{I}$ uses a deep neural network that given an input image $\mathbf{I}_t$, outputs an image of the same size $\costmap$. We refer to $\costmap$ as the image feature \emph{costmap} and ${c_t}_{i,j} = \costmap(i, j)$. 
We use a fully convolutional network with the same architecture as that used by~\citet{zhou2017scene} for image segmentation. The network is composed of the MobileNetV2~\citep{sandler2018mobilenetv2} as the encoder and a transposed convolution layer with deep supervision as the decoder.

\vspace{-0.5em}
\subsubsection{Collection of Training Data} \label{sec:train_data_col}
\vspace{-0.5em}
\begin{figure*}[b]
 \centering
 \includegraphics[width=1\linewidth ,trim=0 0 0 0,clip]{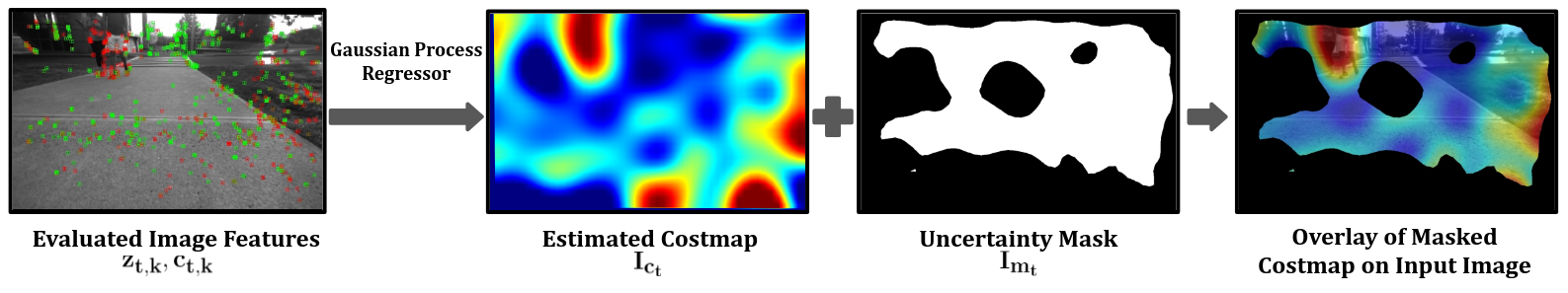}
 \caption{Training label generation for the introspective function.}
 \label{fig:costmap_example}
\end{figure*}

\begin{wrapfigure}[20]{r}{0.5\textwidth}
  \vspace{-3.5em} %
  \resizebox{0.50\textwidth}{!}{%
  \begin{minipage}{0.57\textwidth}
  \begin{algorithm}[H]
  \small
  \caption{\textsc{Training Label Generation}}\label{alg:Train Data Generation}
  \begin{algorithmic}[1]
  \State $\textbf{Input:}$ Set of matched features and map points $\{(\obs_{t,k}, \mathbf{p}_k)\}_{k=1}^{N}$ for current frame, estimated camera pose $\camPoseEst$, reference camera pose $\camPose$, reference camera pose covariance $\Sigma_{\camPose} $
  \State $\textbf{Output:}$ Costmap image $\mathbf{I_c}_t$
  
  \State $l\gets$ costmap computation grid cell size
  \State $(h, w)\gets$ output cost-map size
  \State $\mathbf{I_g}[$\texttt{floor}($\frac{h}{l}$)$][$\texttt{floor}($\frac{w}{l}$)]
  
  \State $\Delta \mathbf{T}_t \gets \hat{\mathbf{T}}_{t}^w ~  \mathbf{T}_w^t$
  \If{\textsc{IsTrackingUnreliable($\Delta \mathbf{T}_t, \Sigma_{\camPose})$}} \
    \State return $-1$.
    
  \EndIf
  
  \For {$k\gets 1$ to $N$}
    \State $\reprojErr_{t,k}\gets$ \textsc{CalcReprojectionError}($\obs_{t,k}, \mathbf{p}_k$)
    \State $c_{t,k}\gets \reprojErr_{t,k}^T \Sigma_{t,k}^{-1} \reprojErr_{t,k}$
  \EndFor

  \For {$i\gets 0$ to \texttt{floor}($\frac{w}{l}$)}
    \For {$j\gets 0$ to \texttt{floor}($\frac{h}{l}$)} 	
      \State $\mathbf{y}\gets$ ($il + l/2, jl + l/2$)
      \State $\mathbf{I_g}[i][j]\gets$ \textsc{GaussianProc}($\mathbf{y}, \{(\obs_{t,k}, c_{t,k}) \}_{k=1}^N$)
      
    \EndFor
  \EndFor
  \State $\mathbf{I_c}_t \gets$ \textsc{ResizeImage}($\mathbf{I_g}, (h, w)$)
  
  \end{algorithmic}
  \end{algorithm}
  \end{minipage}
  }
  \end{wrapfigure}

\thisWork~ requires a set of pairs of input images and their corresponding target image feature costmaps $\mathbf{D} = \{(\mathbf{I}_t, \mathbf{I_c}_t)\}$ to train the introspection function. 
The training is performed offline and loose supervision is provided in the form of estimates of the reference pose of the camera $\{\camPose\}$ by a 3D lidar-based SLAM solution~\cite{shan2018lego}. $\{\camPose\}$ is only used to 
flag the frames, at which the tracking accuracy of V-SLAM is low, so they are not used for training data generation.
The automated procedure for generating the dataset $\mathbf{D}$, presented in Algorithm~\ref{alg:Train Data Generation}, is as follows:
The V-SLAM algorithm is run on the images and at each frame $\mathcal{K}_t$, the Mahalanobis distance of the reference and estimated pose of the camera is calculated as
\begin{align}
d^t &= {\delta \mathbf{T}_t}^{T}  \Sigma^{-1}_{\camPose} \delta \mathbf{T}_t \\
\delta \mathbf{T}_t &= \mathrm{Log} \left(\Delta \mathbf{T}_t \right) = \mathrm{Log} \left(\hat{\mathbf{T}}_{t}^w ~  \mathbf{T}_w^t \right)
\end{align}
where $\delta \mathbf{T}_t \in \mathfrak{se}(3)$ denotes the corresponding element of $\Delta \mathbf{T}_t$ in the Lie algebra of $SE(3)$. $\Sigma_{\camPose}$ is the covariance of the reference pose of the camera and is approximated as a constant lower bound for the covariance of the reference SLAM solution.
A $\chi^2$ test with $\alpha=0.05$ is done for $d^t$ and if it fails, the current frame will be flagged as unreliable and a training label will not be generated for it.
At each frame $\mathcal{K}_t$ recognized as reliable for training data labeling, reprojection error values $\reprojErr_{t, k}$ are calculated for all matched image features. A normalized cost value $c_{t, k} = \reprojErr_{t,k}^T \Sigma_{t,k}^{-1} \reprojErr_{t,k}$ is then computed for each image feature, where $\Sigma_{t,k}$ denotes the diagonal covariance matrix associated
with the scale at which the feature has been extracted.
The set of sparse cost values calculated for each frame is then converted to a costmap $\mathbf{I_c}_t$ the same size as the input image.
This is achieved
using a Gaussian Process regressor.
The generated costmaps along with the input images are then used to train the introspection function using a stochastic gradient descent (SGD) optimizer and a mean squared error loss (MSE) that is only applied to the regions of the image where the uncertainty of the output of the GP is lower than a threshold.
Figure~\ref{fig:costmap_example} shows the estimated costmap and the uncertainty mask for an example input image.

The advantage of such a self-supervised training scheme is that it removes the need for highly accurate ground truth poses of the camera, which would have been necessary if image features were to be evaluated by common approaches such as the epipolar error across different frames. %

\vspace{-0.5em}
\subsection{Robust Estimation in IV-SLAM}
\vspace{-0.5em}

During inference, input images are run through the introspection function $\mathbb{I}$ to obtain the estimated costmaps $\costmapEst$. \thisWork{} uses $\costmapEst$ to both guide the feature extraction process and adjust the contribution of extracted features when solving for $\camPoseAll$ and $\mathbf{M}$.

\vspace{-0.5em}
\paragraph{Guided feature extraction.}
Each image $\mathbf{I}_t$ is divided into equally sized cells and the maximum number of image features to be extracted from each cell $C_k$ is determined to be inversely proportional to $\sum_{(i, j) \in C_k} \costmapEst(i, j)$, i.e. the sum of the corresponding costmap image pixels within that cell. This helps \thisWork{} prevent situations where the majority of extracted features in a frame are unreliable.

\vspace{-0.5em}
\paragraph{Reliability-aware bundle adjustment.}
Extracted features from the input image are matched with map points, and for each matched image feature $\obs_{t,k}$ extracted at pixel location $(i, j)$, a specific loss function $L_{\lossDelta}$ is generated as defined in Eq.~\ref{eq:huber_loss}.
The loss function parameter $\lossDelta \in [0, \delta_{\text{max}}]$ is approximated as $\frac{1 - \hat{c}_{t, k}}{1 + \hat{c}_{t, k}} \delta_{\text{max}}$, where $\hat{c}_{t, k} = \costmapEst(i, j) \in [0, 1]$ and $\delta_{\text{max}}$ is a positive constant and a hyperparameter that defines the range at which $\lossDelta$ can be adjusted. We pick $\delta_{\text{max}}$ to be the $\chi^2$ distribution's $95^{\text{th}}$ percentile, i.e. $7.82$ for a stereo observation.
In other terms, for each image feature, the Huber loss is adjusted such that the features that are predicted to be less reliable (larger $c_{t, k}$) have a less steep associated loss function (smaller $\lossDelta$).
Lastly, the tracked features along with their loss functions are plugged into Eq.~\ref{eq:bundle_adjustment} and the solution to the bundle adjustment problem, i.e. the current pose of the camera as well as adjustments to the previously estimated poses and the location of map points, are estimated using a nonlinear optimizer.

%% file: experimental_results.tex
\section{Experimental Results} \label{experimental_results}
In this section:
\begin{inparaenum}[1)]
\item We evaluate \thisWork{} on how well it predicts reliability of image features (Section~\ref{sec:feature_reliability}).
\item We show that \thisWork{} improves tracking accuracy of a state-of-the-art V-SLAM algorithm and reduces frequency of its tracking failures (Section~\ref{sec:tracking_acc}).
\item We look at samples of sources of failure learned by \thisWork{} to negatively affect V-SLAM. (Section~\ref{sec:qual_results}).
\end{inparaenum}
To evaluate \thisWork{}, we implement it on top of the stereo version of ORB-SLAM~\citep{mur2017orb}. We pick ORB-SLAM because it has various levels of feature matching pruning and outlier rejection in
place, which indicates that the remaining failure cases that we address with introspective vision cannot be resolved with meticulously engineered outlier rejection methods.

\vspace{-0.5em}
\subsection{Experimental Setup}\label{sec:experimental_setup}
\vspace{-0.5em}
State-of-the-art vision-based SLAM algorithms have shown great performance on popular benchmark datasets such as \citep{geiger2012we} and EuRoC~\citep{burri2016euroc}. These datasets, however, do not reflect the many difficult situations that can happen when the robots are deployed in the wild and over extended periods of time~\citep{shi2019we}. In order to assess the effectiveness of \thisWork{} on improving visual SLAM performance, in addition to evaluation on the public EuRoC and KITTI datasets, we also put IV-SLAM to test on simulated and real-world datasets that we have collected to expose the algorithm to challenging situations such as reflections, glare, shadows, and dynamic objects.

\vspace{-0.5em}
\paragraph{Simulation.}
In order to evaluate \thisWork{} in a controlled setting, where we have accurate poses of the camera and ground truth depth of the scene, we use AirSim~\citep{shah2018airsim}, a photo-realistic simulation environment. A car is equipped with a stereo pair of RGB cameras as well as a depth camera that provides ground truth depth readings for every pixel in the left camera frame. The dataset encompasses more than \SI{60}{\kilo\meter} traversed by the car in different environmental and weather conditions such as clear weather, wet roads, and in the presence of snow and leaves accumulation on the road.

\vspace{-0.5em}
\paragraph{Real-world.} 
We also evaluate \thisWork{} on real-world data that we collect using a Clearpath Jackal robot equipped with a stereo pair of RGB cameras as well as a Velodyne VLP-16 3D Lidar. The dataset consists of more than \SI{7}{\kilo\meter} worth of trajectories traversed by the robot over the span of a week in a college campus setting in both indoor and outdoor environments and different lighting conditions.
The reference pose of the robot is estimated by a 3D Lidar-based SLAM solution, LeGO-LOAM~\citep{shan2018lego}. The algorithm is run on the data offline and with extended optimization rounds for increased accuracy.
For both the real-world and simulation datasets the data is split into training and test sets, each composed of separate full deployment sessions(uninterrupted trajectories), such that both training and test sets include data from all environmental conditions.

\subsection{Feature Reliability Prediction} \label{sec:feature_reliability}

\begin{wrapfigure}[14]{r}{5.5cm}
\vspace{-4em} %
\includegraphics[width=5.5cm]{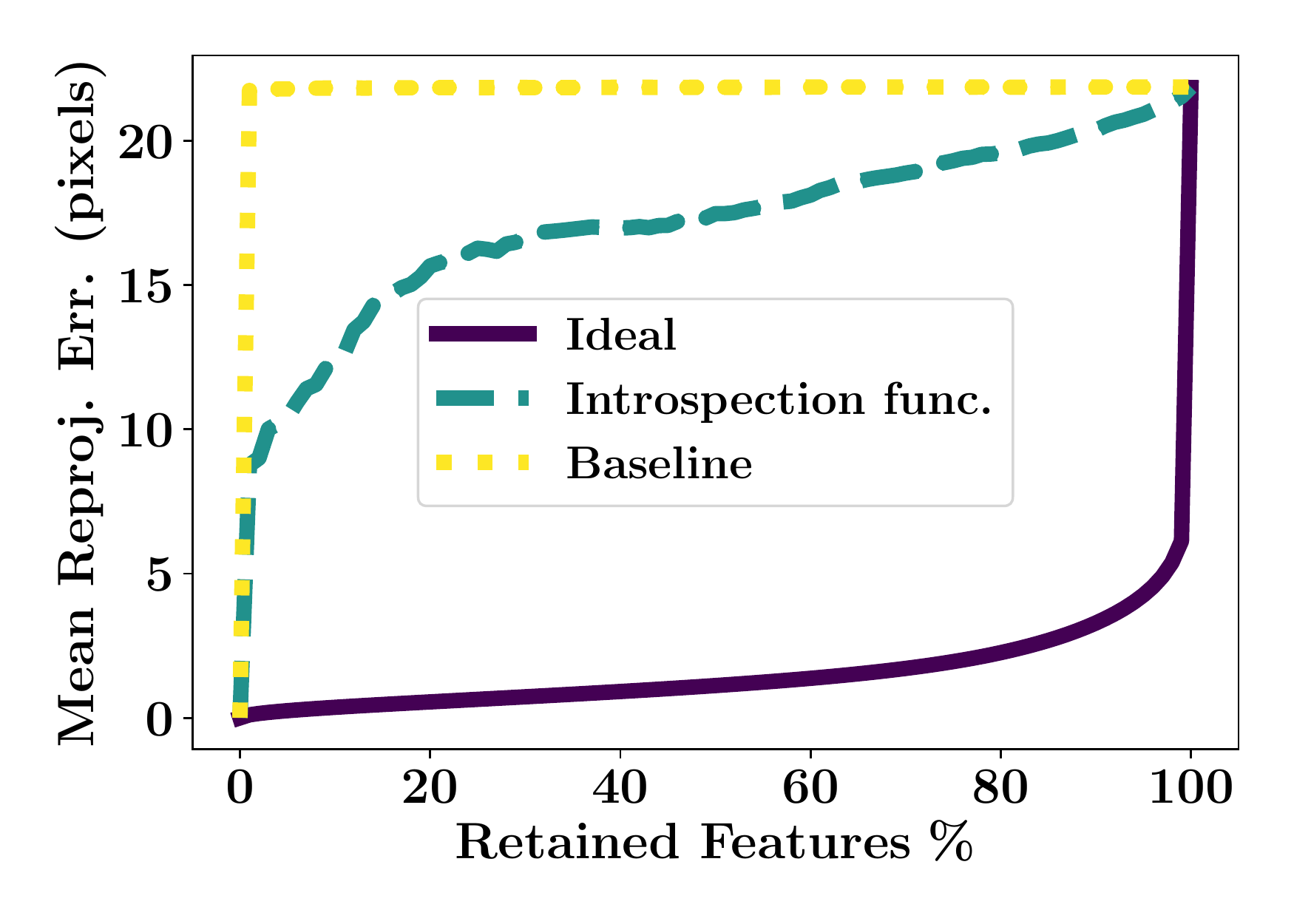}
\caption{Mean reprojection error for image features sorted
1) randomly (baseline),
2) based on predicted reliability (Introspection func.),
3) based on ground truth reprojection error (ideal)
}\label{fig:feature_qual_pred}
\end{wrapfigure}

In order to evaluate \thisWork's introspection function (IF) on predicting reliability of image features, we compare IF's predicted reliability of image features with their corresponding ground truth reprojection error. Since obtaining the ground truth reprojection error requires access to ground truth 3D coordinates of objects associated with each image feature as well as accurate reference poses of the camera, we conduct this experiment in simulation.
\thisWork{} is trained on the simulation dataset with the method explained in Section~\ref{method}. The IF is then run on all images in the test set along with the original ORB-SLAM. For each image feature extracted and matched by ORB-SLAM, we log its predicted cost by the IF, as well as its corresponding ground truth reprojection error.
We then sort all image features in ascending order with respect to \begin{inparaenum}[1)]
\item ground truth reprojection errors and
\item predicted cost values.
\end{inparaenum}
Figure~\ref{fig:feature_qual_pred} illustrates the mean reprojection error for the top $x\%$ of features in each of these ordered lists for a variable $x$. The lower the area under a curve in this figure, the more accurate is the corresponding image feature evaluator in sorting features based on their reliability. The curve corresponding to the ground truth reprojection errors indicates the ideal scenario where all image features are sorted correctly. The baseline is also illustrated in the figure as the resultant mean reprojection error when the features are sorted randomly (mean error over 1000 trials) and it corresponds to the case when no introspection is available and all features are treated equally. As can be seen, using the IF significantly improves image feature reliability assessment.

\setlength\tabcolsep{1.0pt}

\begin{table}[]
  \parbox{.45\linewidth}{
  \caption{\textsc{Tracking Accuracy in the KITTI Dataset}}
  \label{table:kitti_results}
  \centering
  \resizebox{0.5\textwidth}{!}{
  \begin{tabular}{l c c c c }
    \toprule
    \multicolumn{1}{l}{} &
  	\multicolumn{2}{c}{IV-SLAM} &
  	\multicolumn{2}{c}{ORB-SLAM} \\
    \cmidrule(r{0.7em}){2-3}
	\cmidrule(l{0.0em}){4-5}    
    
  	\multirow[b]{2}{1.25cm}{Sequence} &
 	\multirow[b]{2}{1.5cm}{\centering Trans. Err. \%} &
 	\multirow[b]{2}{2.0cm}{\centering Rot. Err. (\si{deg/100m})} &
 	\multirow[b]{2}{1.5cm}{\centering Trans. Err. \%} &
 	\multirow[b]{2}{2.0cm}{\centering Rot. Err. (\si{deg/100m})} \\
 	& & & &  \\
    \midrule
 00  & 0.69 		   & 0.25		     & 0.69	                & 0.25  \\
 01  & \textbf{1.43}   & 0.22 		     & 1.47 		        & 0.22  \\
 02  & 0.79		       & \textbf{0.22} 	 & \textbf{0.76}	    & 0.24  \\
 03  & 0.74		       & \textbf{0.19}   & \textbf{0.70} 	    & 0.23  \\
 04  & \textbf{0.49}   & 0.13 		     & 0.55			        & 0.13 \\
 05  & 0.40 		   & 0.16	         & \textbf{0.38}	    & 0.16  \\
 06  & \textbf{0.49}   & \textbf{0.14}   & 0.56				    & 0.19  \\
 07  & 0.49            & \textbf{0.27}   & 0.49				    & 0.29 \\
 08  & \textbf{1.02}   & 0.31 	         & 1.05  				& 0.31  \\
 09  & 0.85 		   & 0.25	         & \textbf{0.82} 		& 0.25 \\
 10  & \textbf{0.61}   & \textbf{0.26}   & 0.62 				& 0.29 \\
    \midrule
 Average  & 0.77       & \textbf{0.24} & 0.77 		        & 0.25 \\
    \bottomrule
  \end{tabular}
    }
  } \qquad
  \parbox{.45\linewidth}{
    \centering
    \caption{\textsc{Tracking Accuracy in the EuRoC Dataset}}
    \label{table:euroc_results}
    \resizebox{0.5\textwidth}{!}{
    \begin{tabular}{l c c c c }
      \toprule
      \multicolumn{1}{l}{} &
      \multicolumn{2}{c}{IV-SLAM} &
      \multicolumn{2}{c}{ORB-SLAM} \\
      \cmidrule(r{0.7em}){2-3}
    \cmidrule(l{0.0em}){4-5}    
      
      \multirow[b]{2}{1.25cm}{Sequence} &
     \multirow[b]{2}{1.5cm}{\centering Trans. Err. \%} &
     \multirow[b]{2}{2.0cm}{\centering Rot. Err. (\si{deg/m})} &
     \multirow[b]{2}{1.5cm}{\centering Trans. Err. \%} &
     \multirow[b]{2}{2.0cm}{\centering Rot. Err. (\si{deg/m})} \\
     & & & &  \\
      \midrule
   MH1  & \textbf{2.26} 		   & \textbf{0.19}		              & 2.42	    & 0.21  \\
   MH2  & 1.78                   & 0.18 		     & \textbf{1.49} 		    & \textbf{0.16} \\
   MH3  & 3.27		           & 0.18 	         & 3.27	                   & \textbf{0.17}  \\
   MH4  & 3.85		           & 0.16            & \textbf{3.49} 	          & \textbf{0.15}  \\
   MH5  & \textbf{2.98}          & \textbf{0.16} 		     & 3.32			               & 0.18 \\
   V1\_1  & 8.93 		           & \textbf{1.05}	         & \textbf{8.85}	             & 1.06  \\
   V1\_2  & \textbf{4.38}         & 0.41            & 4.46				    & \textbf{0.39}  \\
   V1\_3  & \textbf{7.85}         & \textbf{1.24}            & 14.86			    & 2.35 \\
   V2\_1  & \textbf{2.92}         & \textbf{0.76}	         & 4.37  				& 1.39  \\
   V2\_2  & 2.89 		            & 0.62	         & 2.76 		        & \textbf{0.59} \\
   V2\_3  & \textbf{11.00}         & 2.49            & 12.73 				& \textbf{2.39} \\
      \midrule
   Average  & \textbf{4.74}       & \textbf{0.68}            & 5.64 		        & 0.82 \\
      \bottomrule
    \end{tabular}
    }
  }
\end{table}

\subsection{Tracking Accuracy and Tracking Failures} \label{sec:tracking_acc}
We compare our introspection-enabled version of ORB-SLAM with the original algorithm in terms of their camera pose tracking accuracy and robustness. Both algorithms are run on the test data and their estimated poses of the camera are recorded.
If the algorithms loose track due to lack of sufficient feature matches across frames, tracking is reinitialized and continued from after the point of failure along the trajectory and the event is logged as an instance of tracking failure for the corresponding SLAM algorithm.
The relative pose error (RPE) is then calculated for both algorithms at consecutive pairs of frames that are $d$ meters apart as defined in~\cite{schubert2018tum}. 

\mypara{KITTI and EuRoC datasets}
We perform leave-one-out cross-validation separately on the KITTI and EuRoC datasets, i.e. to test IV-SLAM on each sequence, we train it on all other sequences in the same dataset. 
Tables~\ref{table:kitti_results} and \ref{table:euroc_results} compare the per trajectory root-mean-square error (RMSE) of the rotation and translation parts of the RPE for IV-SLAM and ORB-SLAM in the KITTI and EuRoC datasets, respectively.
The baseline ORB-SLAM does a good job of tracking in the KITTI dataset. There exists no tracking failures and the overall relative translational error is less than $1\%$. Given the lack of challenging scenes in this dataset, IV-SLAM performs similar to ORB-SLAM with only marginal improvement in the overall rotational error. While EuRoC is more challenging than the KITTI given the higher mean angular velocity of the robot, the only tracking failure happens in the V2\_3 sequence and similarly for both ORB-SLAM and IV-SLAM due to severe motion blur. IV-SLAM performs similar to ORB-SLAM on the easier trajectories, however, it significantly reduces the tracking error on the more challenging trajectories such as V1\_3. Over all the sequences, IV-SLAM improves both the translational and rotational tracking accuracy.

\setlength\tabcolsep{1.0pt}

\begin{table}[t]
  \caption{\textsc{Aggregate Results for Simulation and Real-world Experiments}}
  \label{table:aggregate_results}
  \centering
  \resizebox{\textwidth}{!}{%
  \begin{tabular}{l c c c c c c }
    \toprule
    \multicolumn{1}{l}{} &
  	\multicolumn{3}{c}{Real-World} &
  	\multicolumn{3}{c}{Simulation} \\
    \cmidrule(r{0.7em}){2-4}
	\cmidrule(l{0.0em}){5-7}    
    
  	\multirow[b]{1}{2.0cm}{Method} &
 	\multirow[b]{1}{3.0cm}{\centering Trans. Err. \%} &
 	\multirow[b]{1}{3.0cm}{\centering Rot. Err. (\si{deg/m})} &
 	\multirow[b]{1}{2.0cm}{\centering MDBF (\si{m})} &
 	\multirow[b]{1}{3.0cm}{\centering Trans. Err. \%} &
 	\multirow[b]{1}{3.0cm}{\centering Rot. Err. (\si{deg/m})} &
    \multirow[b]{1}{2.0cm}{\centering MDBF (\si{m})} \\
    \midrule
 IV-SLAM  & \textbf{5.85} & \textbf{0.511} & \textbf{621.1} & \textbf{12.25} & \textbf{0.172} & \textbf{450.4} \\
 ORB-SLAM  & 9.20 & 0.555 & 357.1 & 18.20 & 0.197 & 312.7 \\
    \bottomrule
  \end{tabular}}
\end{table}

\begin{figure}[t]
 \begin{subfigure}[b]{0.33\linewidth}
   \includegraphics[width=47mm, trim=0 0 0 0,clip]{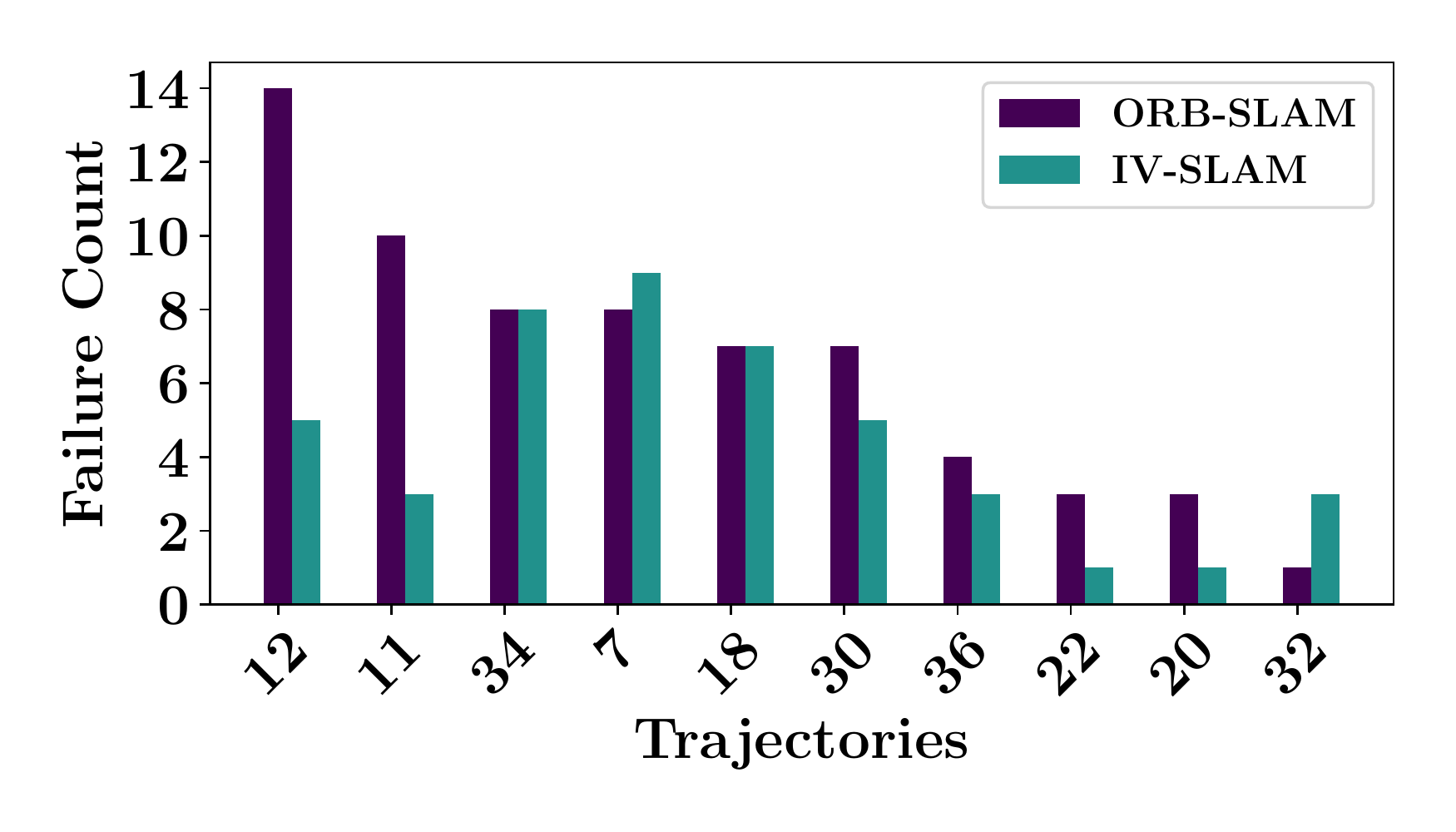}
 	\caption{}
 	\label{fig:failure_count_airsim}
 \end{subfigure}
 \begin{subfigure}[b]{0.33\linewidth}
   \includegraphics[width=47mm, trim=0 0 0 0,clip]{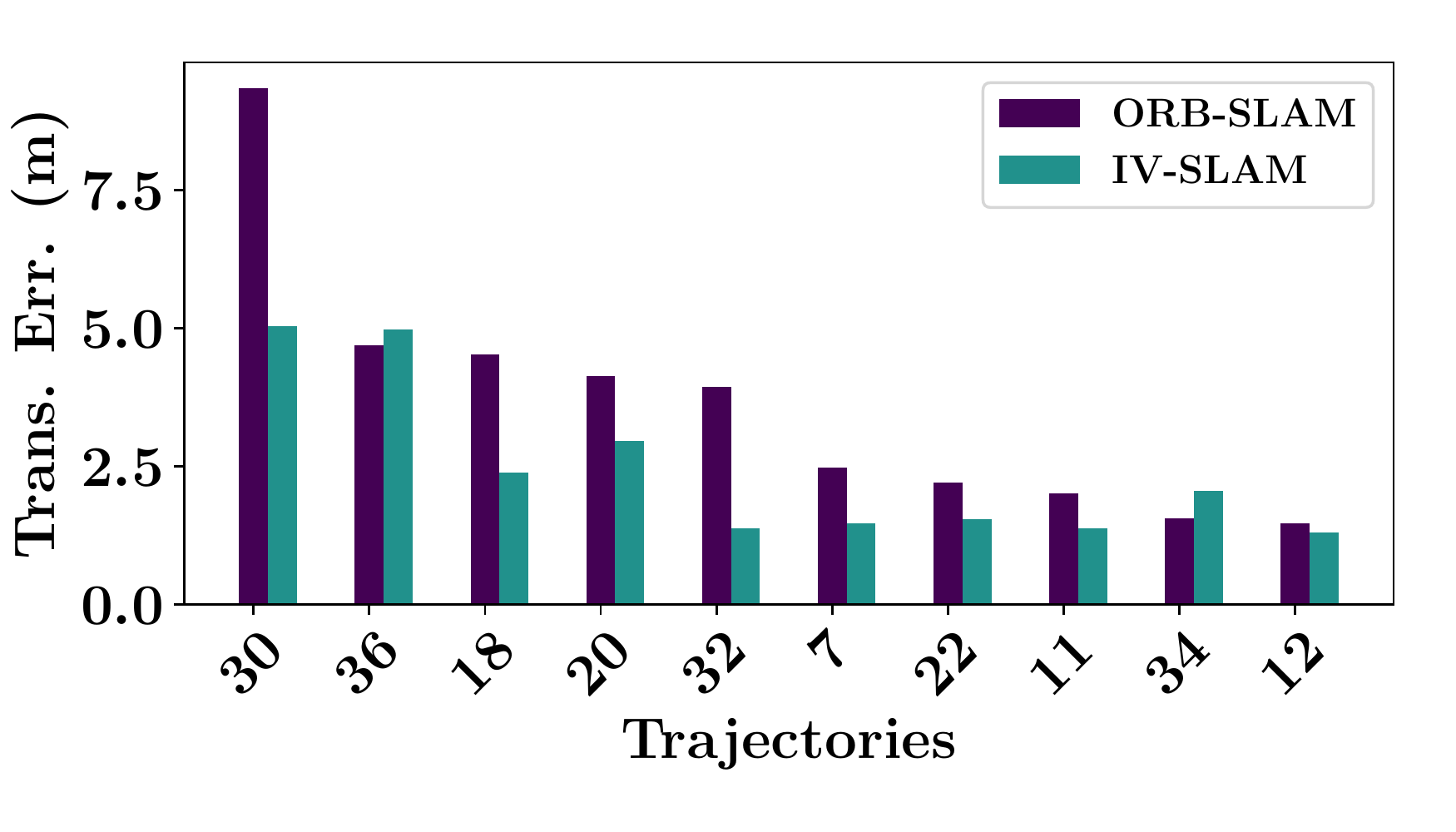}
 	\caption{}
 	\label{fig:rpe_trans_airsim}
 \end{subfigure}
 \begin{subfigure}[b]{0.33\linewidth}
   \includegraphics[width=47mm, trim=0 0 0 0,clip]{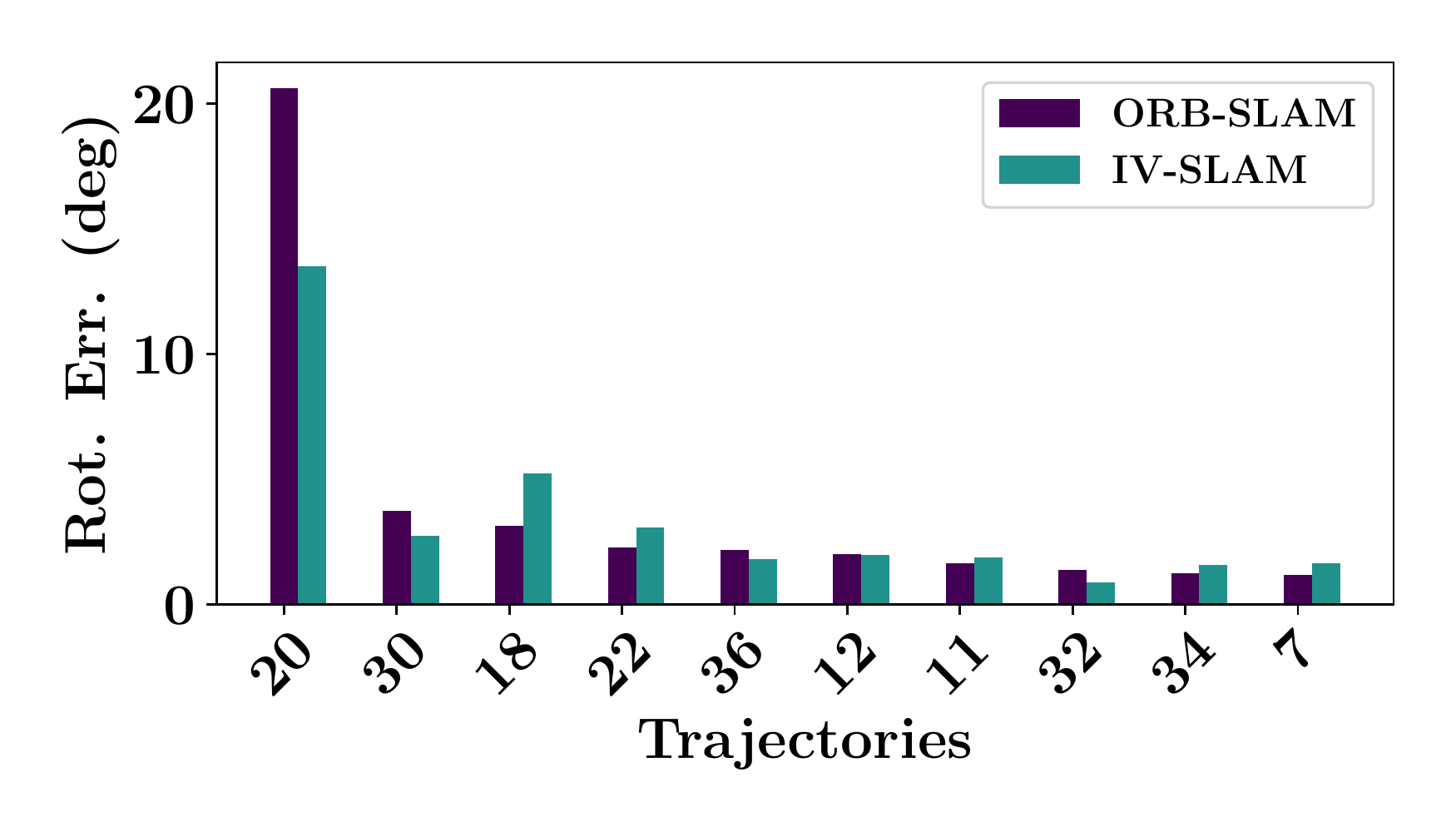}
 	\caption{}
 	\label{fig:rpe_rot_airsim}
 \end{subfigure}
 \caption{Per trajectory comparison of the performance of \thisWork{} and ORB-SLAM in the simulation experiment. (\subref{fig:failure_count_airsim}) Tracking failure count. (\subref{fig:rpe_trans_airsim}) RMSE of translational error and (\subref{fig:rpe_rot_airsim}) RMSE of rotational error over consecutive $20$\si{m}-long horizons.}
 \label{fig:per_traj_results_airsim}
 \vspace{-3mm}
\end{figure}

\begin{figure}[t]
 \begin{subfigure}[b]{0.33\linewidth}
   \includegraphics[width=47mm, trim=0 0 0 0,clip]{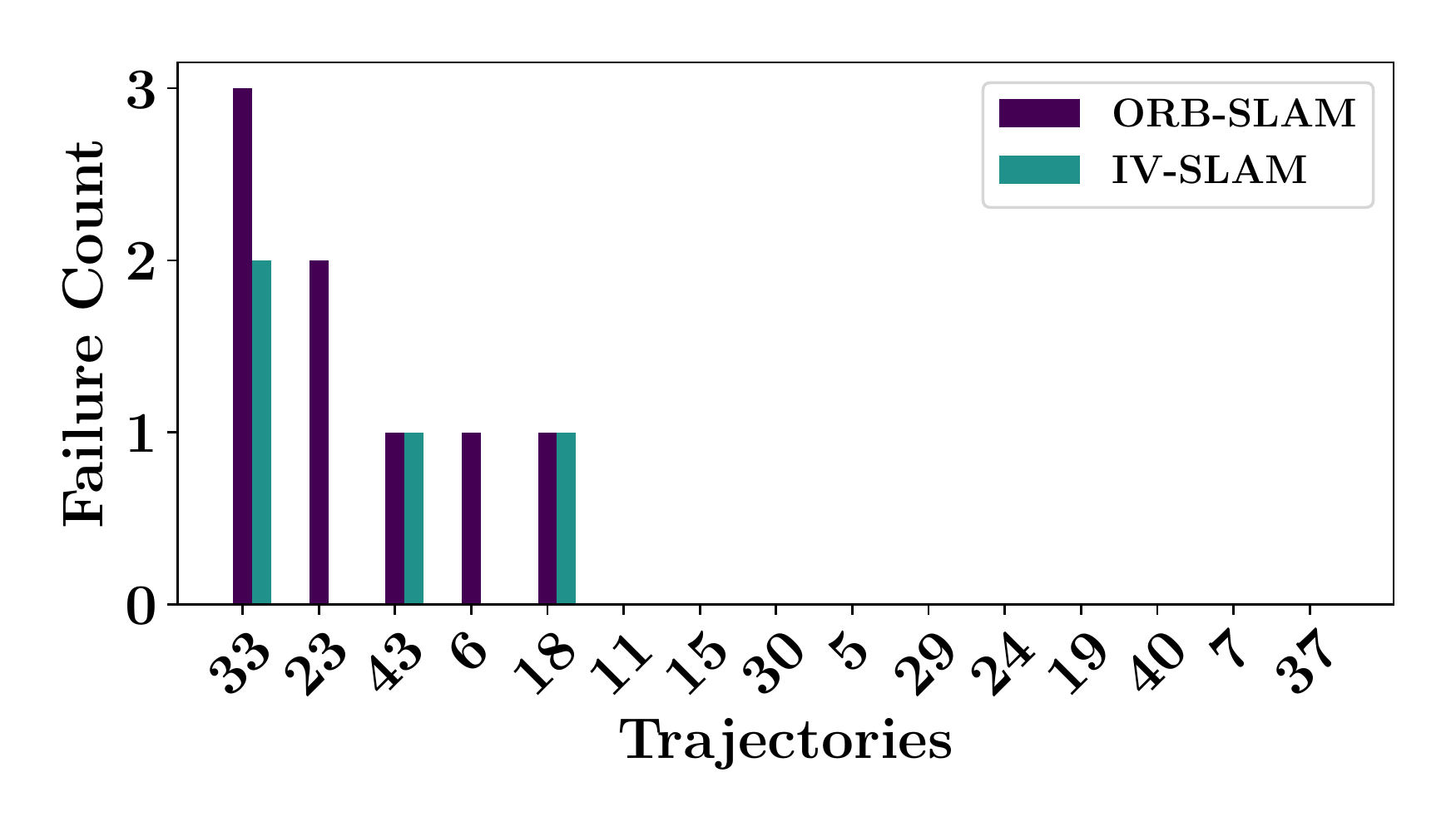}
 	\caption{}
 	\label{fig:failure_count_jackal}
 \end{subfigure}
 \begin{subfigure}[b]{0.33\linewidth}
   \includegraphics[width=47mm, trim=0 0 0 0,clip]{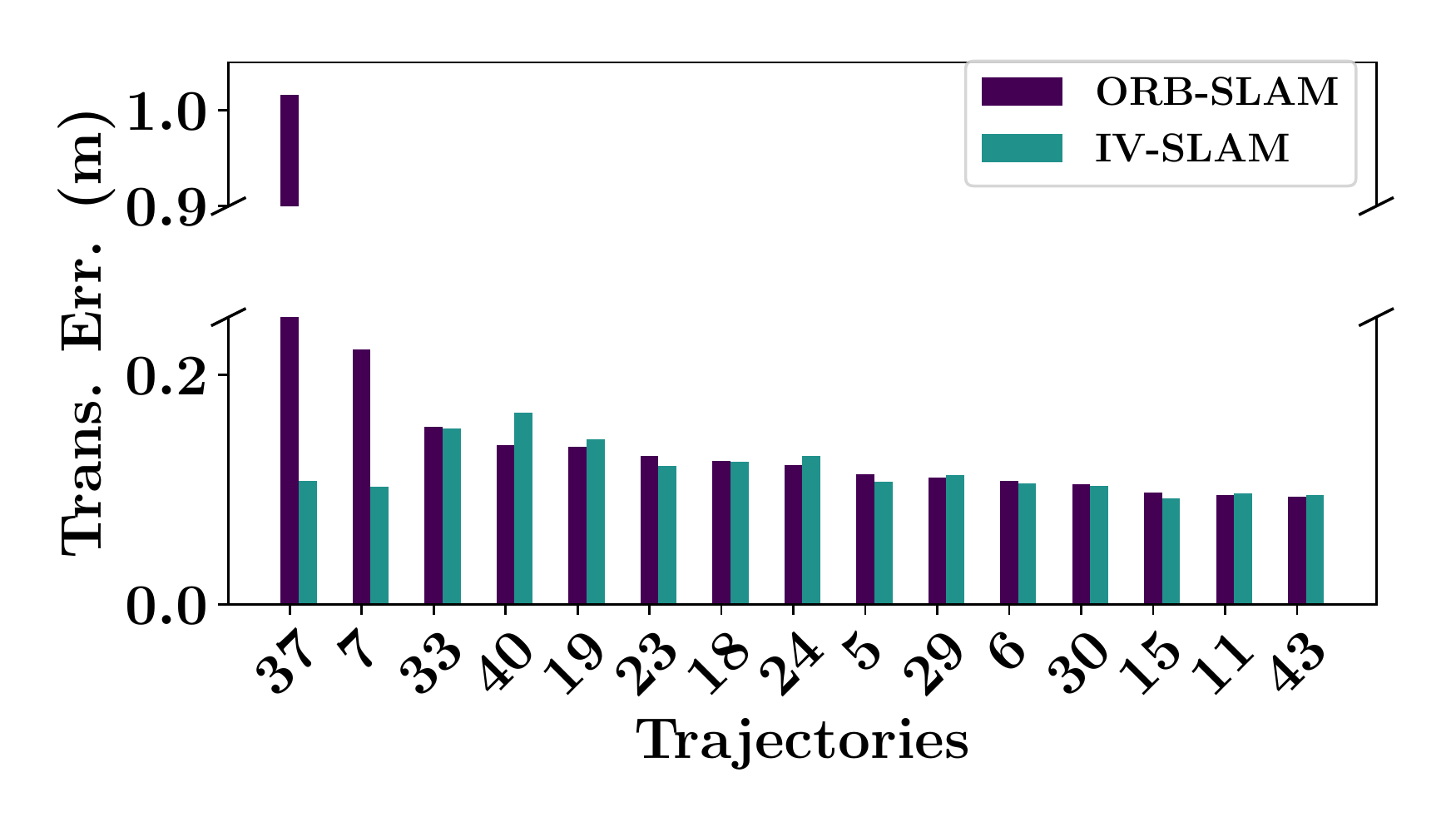}
 	\caption{}
 	\label{fig:rpe_trans_jackal}
 \end{subfigure}
 \begin{subfigure}[b]{0.33\linewidth}
   \includegraphics[width=47mm, trim=0 0 0 0,clip]{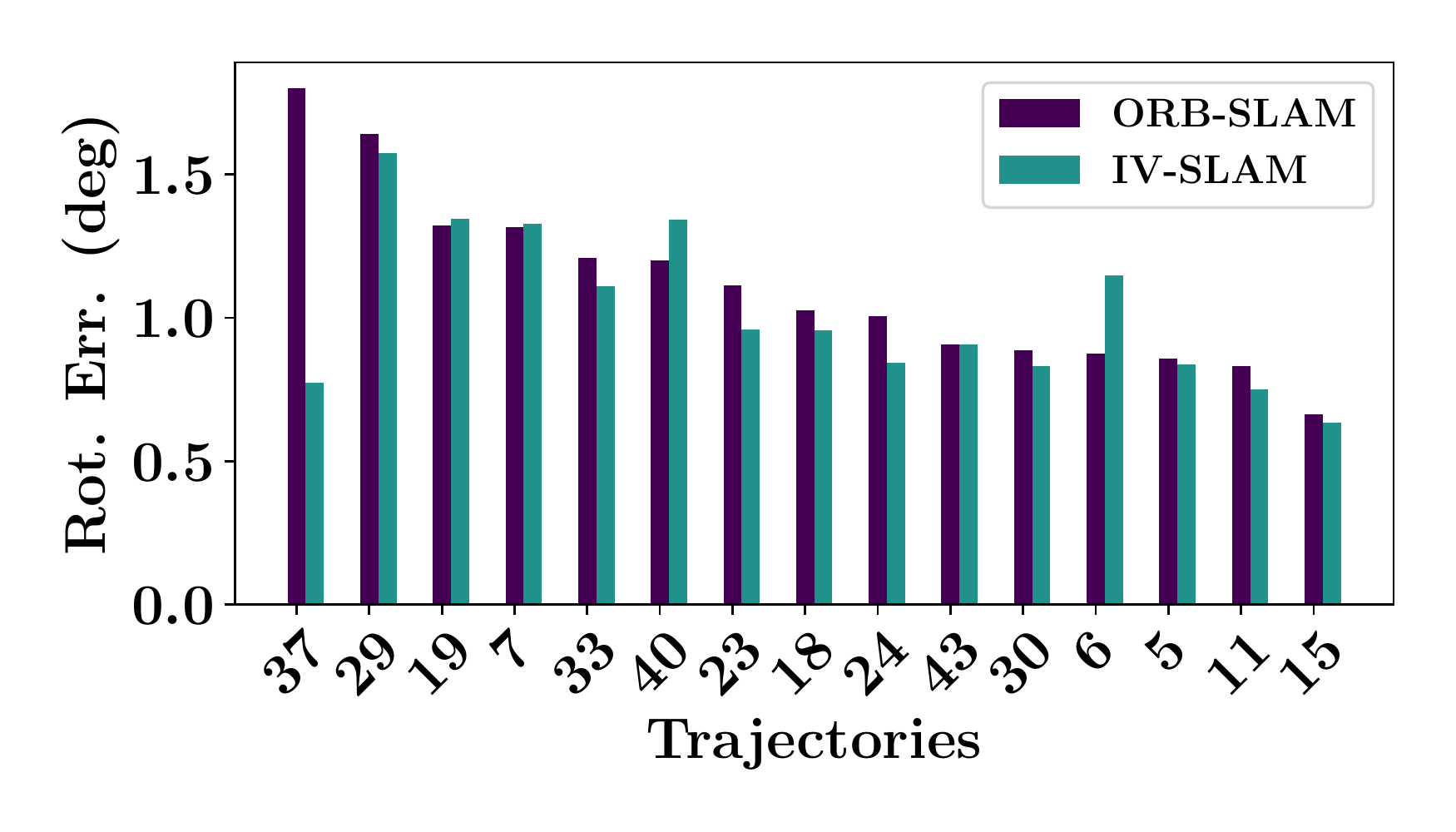}
 	\caption{}
 	\label{fig:rpe_rot_jackal}
 \end{subfigure}
 \caption{Per trajectory comparison of the performance of \thisWork{} and ORB-SLAM in the real-world experiment. (\subref{fig:failure_count_jackal}) Tracking failure count. (\subref{fig:rpe_trans_jackal}) RMSE of translational error and (\subref{fig:rpe_rot_jackal}) RMSE of rotational error over consecutive $2$\si{m}-long horizons.}
 \label{fig:per_traj_results_jackal}
 \vspace{-3mm}
\end{figure}

\mypara{Challenging datasets}
We also evaluate IV-SLAM on the simulation and real-world datasets introduced in Section~\ref{sec:experimental_setup}, which include scenes that are representative of the challenging scenarios that can happen in the real-world applications of V-SLAM.
Given the larger scale of the environment, and faster speed of the robot in the simulation dataset, we pick $d_R=2$\si{m} for the real-world environment and $d_S=20$\si{m} for the simulation. Figures~\ref{fig:per_traj_results_airsim} and~\ref{fig:per_traj_results_jackal} compare the per trajectory tracking error values as well as the tracking failure count for \thisWork{} and ORB-SLAM in both experimental environments. Table~\ref{table:aggregate_results} summarizes the results and shows the RMSE values calculated over all trajectories.
The results show that \thisWork{} leads to more than $70\%$ increase in the mean distance between failures (MDBF) and a $35\%$ decrease in the translation error in the real-world dataset. \thisWork{} similarly outperforms the original ORB-SLAM in the simulation dataset by both reducing the tracking error and increasing MDBF.
As it can be seen numerous tracking failures happen in both environments and the overall error rates are larger than those corresponding to the KITTI and EuRoC datasets due to the more difficult nature of these datasets. It is noteworthy that the benefit gained from using IV-SLAM is also more pronounced on these datasets with challenging visual settings.

\subsection{Qualitative Results} \label{sec:qual_results}
\begin{figure}[b]
 \begin{subfigure}[b]{0.195\linewidth}
   \includegraphics[width=29mm, trim=0 0 0 0,clip]{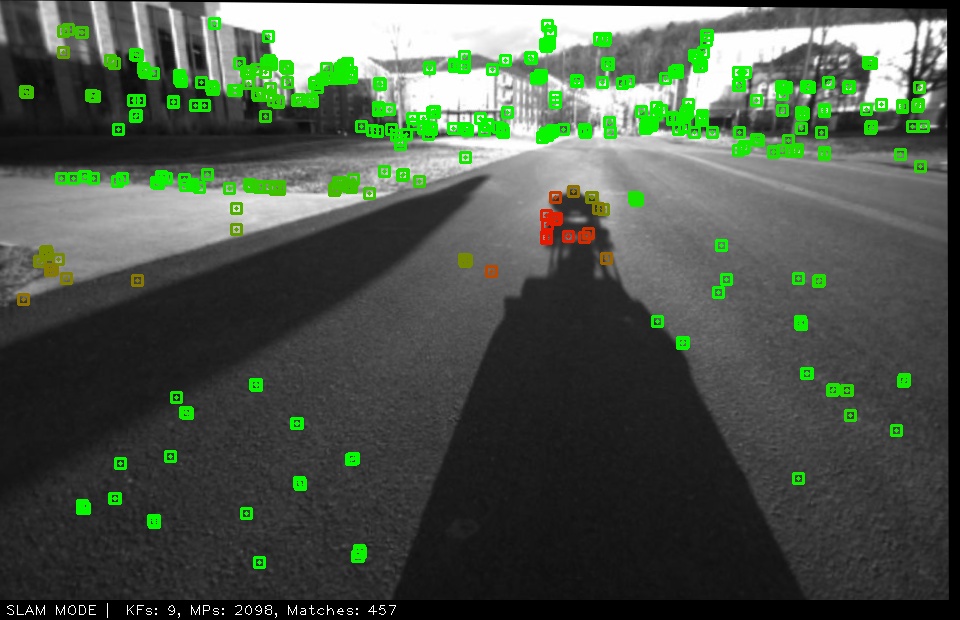}
 \end{subfigure} 
 \begin{subfigure}[b]{0.195\linewidth}
   \includegraphics[width=29mm, trim=0 0 0 0,clip]{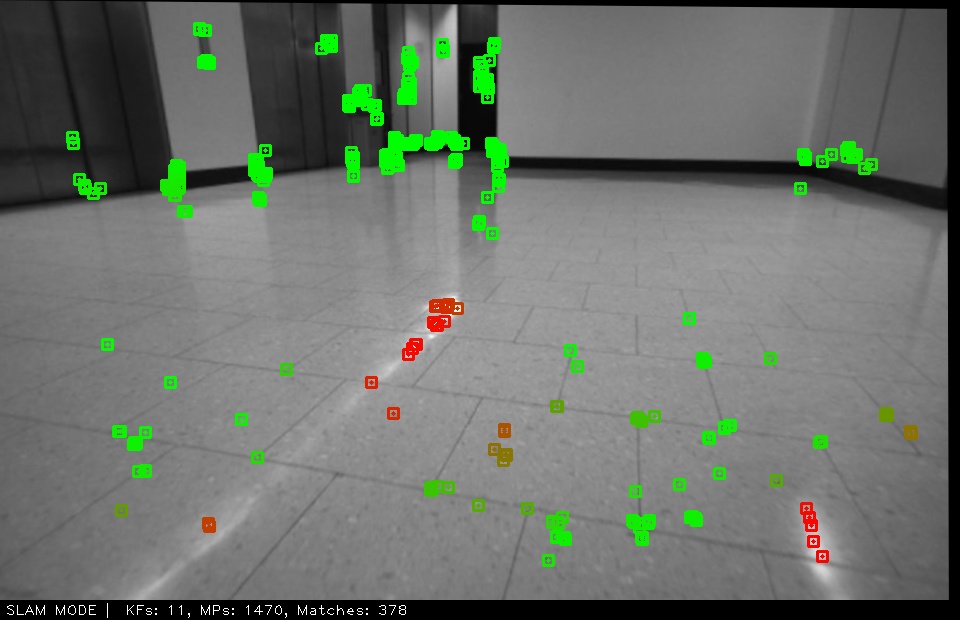}
 \end{subfigure} 
 \begin{subfigure}[b]{0.195\linewidth}
   \includegraphics[width=29mm, trim=0 0 0 0,clip]{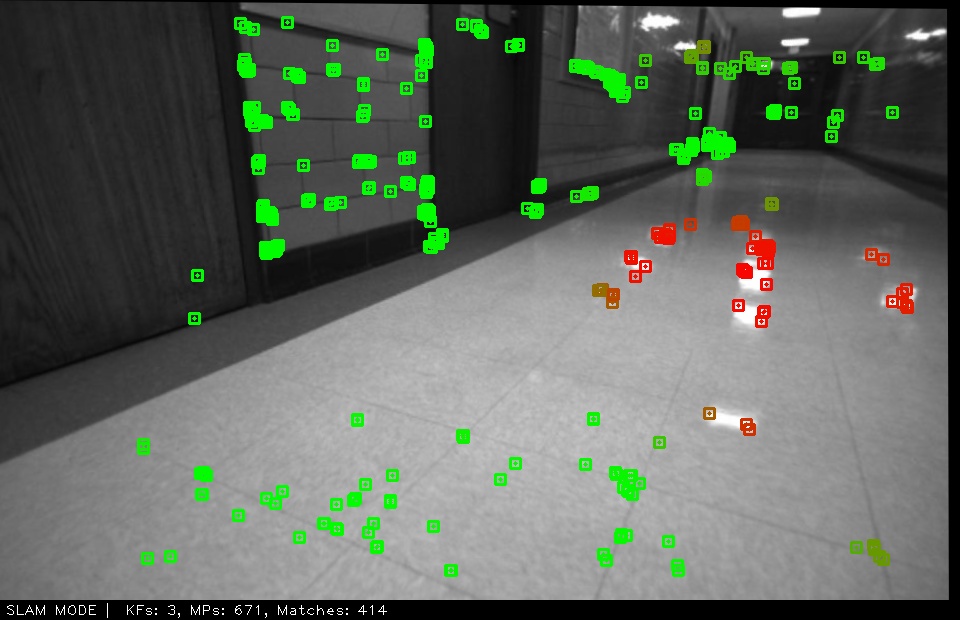}
 \end{subfigure} 
 \begin{subfigure}[b]{0.195\linewidth}
   \includegraphics[width=29mm, trim=0 0 0 0,clip]{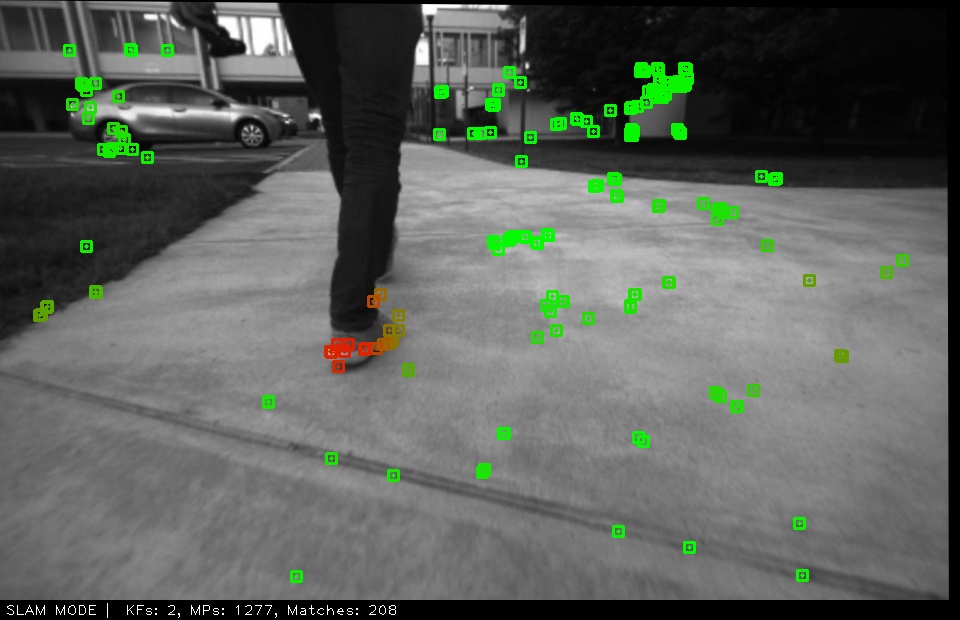}
 \end{subfigure}
 \begin{subfigure}[b]{0.195\linewidth}
   \includegraphics[width=29mm, trim=0 0 0 0,clip]{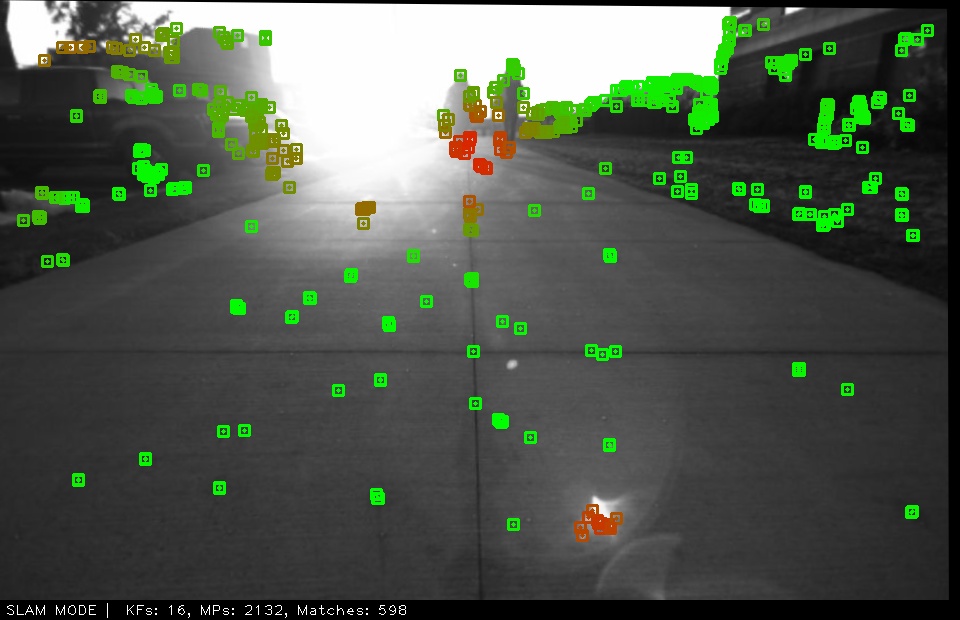}
 \end{subfigure}
  \begin{subfigure}[b]{0.195\linewidth}
   \includegraphics[width=29mm, trim=0 0 0 0,clip]{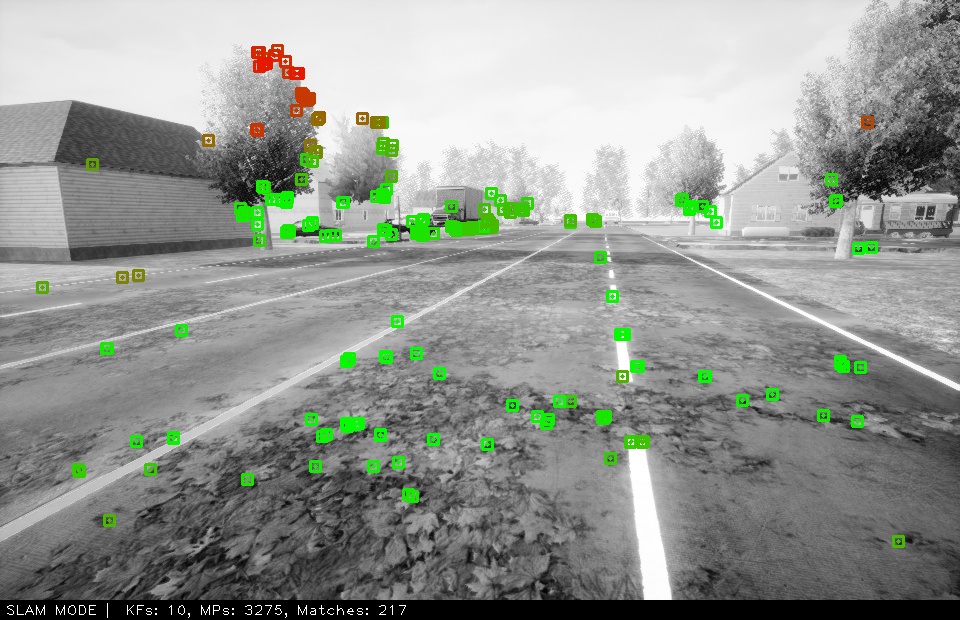}
 \end{subfigure} 
 \begin{subfigure}[b]{0.195\linewidth}
   \includegraphics[width=29mm, trim=0 0 0 0,clip]{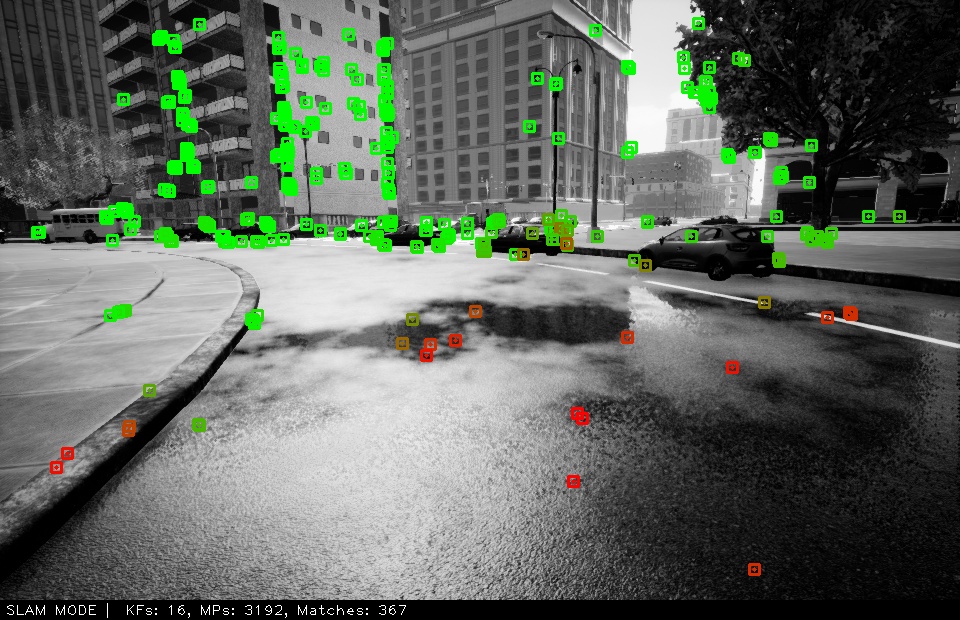}
 \end{subfigure} 
 \begin{subfigure}[b]{0.195\linewidth}
   \includegraphics[width=29mm, trim=0 0 0 0,clip]{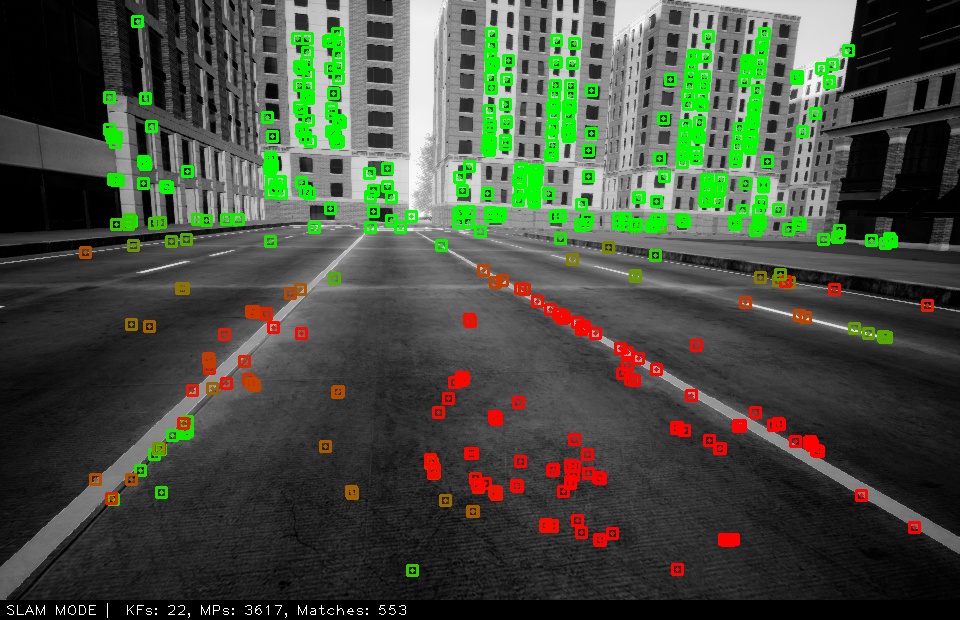}
 \end{subfigure} 
 \begin{subfigure}[b]{0.195\linewidth}
   \includegraphics[width=29mm, trim=0 0 0 0,clip]{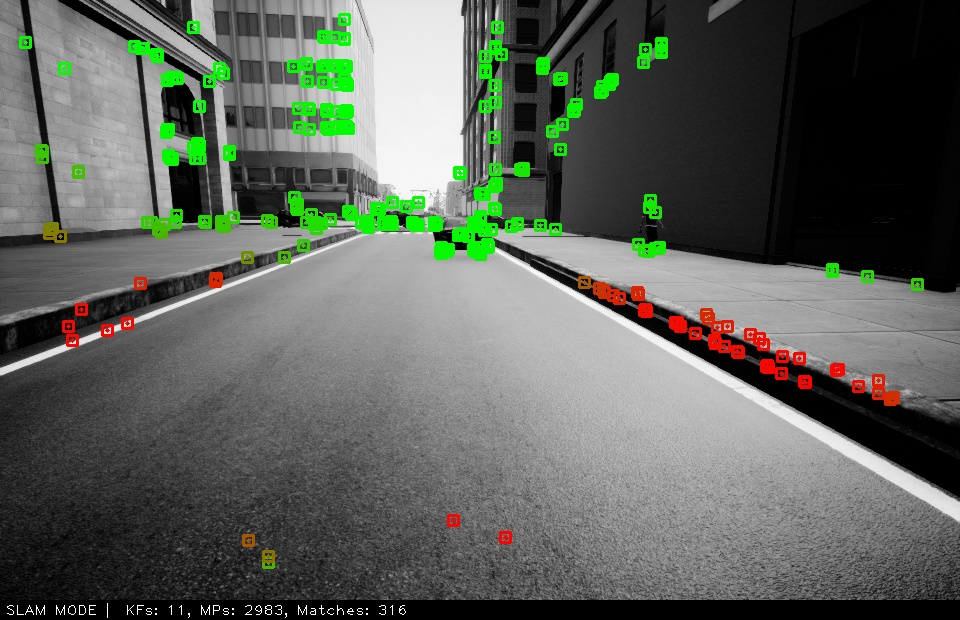}
 \end{subfigure}
 \begin{subfigure}[b]{0.195\linewidth}
   \includegraphics[width=29mm, trim=0 0 0 0,clip]{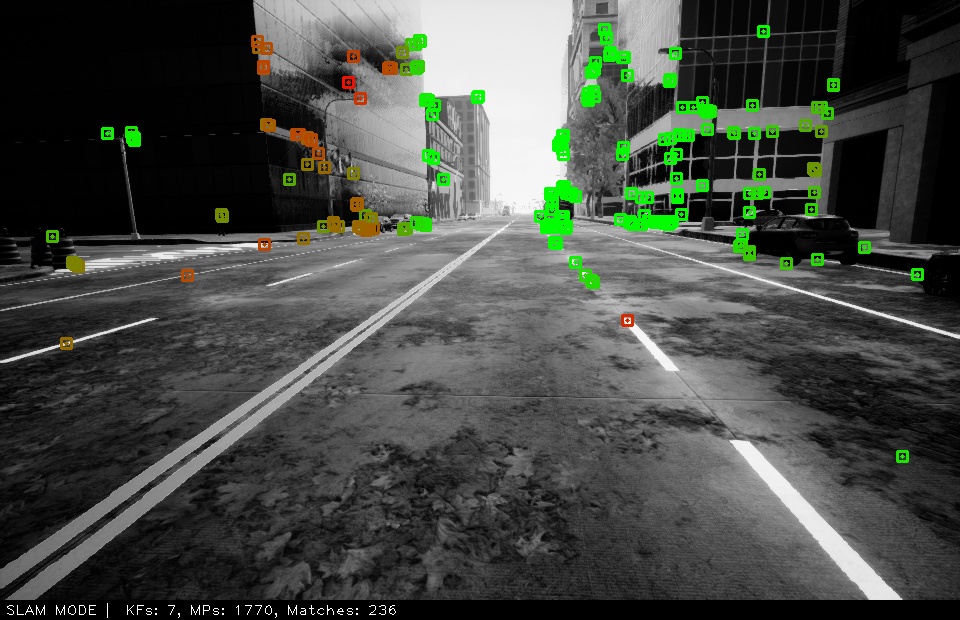}
 \end{subfigure}

 \caption{Snapshots of \thisWork~running on real-world data (top row) and in simulation (bottom row). Green and red points on the image represent the reliable and unreliable tracked image features, respectively, as predicted by the introspection model. Detected sources of failure include shadow of the robot, surface reflections, pedestrians, glare, and ambiguous image features extracted from high-frequency textures such as asphalt or vegetation.}
 \label{fig:image_snapshots}
 \vspace{-3mm}
\end{figure}

In order to better understand how \thisWork{} improves upon the underlying SLAM algorithm, and what it has learned to achieve the improved performance, we look at sample qualitative results.
Figure~\ref{fig:trajectory_example_jackal} demonstrates example deployment sessions of the robot from the real-world dataset and compares the reference pose of the camera with the estimated trajectories by both algorithms under test. It shows how image features extracted from the shadow of the robot and 
surface reflections cause significant tracking errors for ORB-SLAM, while \thisWork{} successfully handles such challenging situations. 
Figure~\ref{fig:image_snapshots} illustrates further potential sources 
of failure picked up by \thisWork{} during inference,
and demonstrates that the algorithm has learned to down-weight image features extracted from sources such as shadow of the robot, surface reflections, lens flare, and pedestrians in order to achieve improved performance.

%% file: conclusion.tex
\vspace{0em}
\section{Conclusion} \label{conclusion}
\vspace{0em}
\begin{wrapfigure}[16]{b}{6.0cm}
  \vspace{-4em} %
  \includegraphics[width=6.0cm]{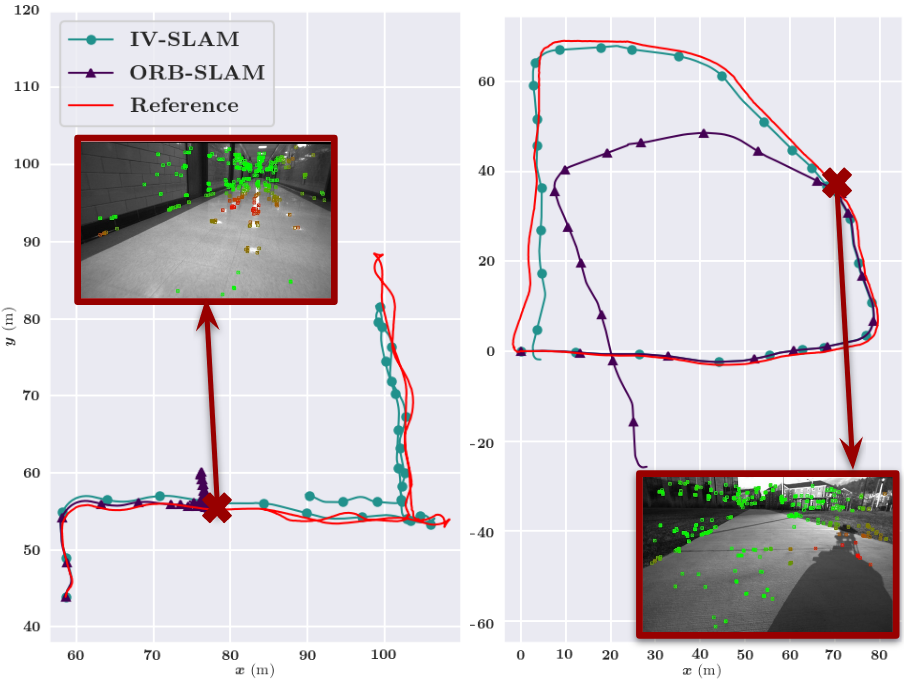}
  \caption{Example deployment sessions of the robot. \thisWork{} successfully follows the reference camera trajectory while ORB-SLAM leads to severe tracking errors caused by image features extracted on the shadow of the robot and surface reflections.}
   \label{fig:trajectory_example_jackal}
\end{wrapfigure}
In this paper, we introduced \thisWork{}, a self-supervised approach for learning to predict sources of failure for V-SLAM and to estimate a context-aware noise model for image correspondences. We empirically demonstrated that~\thisWork{} improves the accuracy and robustness of a state-of-the-art V-SLAM solution with extensive simulated and real-world data.
\thisWork{} currently only uses static input images to predict the reliability of features. 
As future work, we would like to expand the architecture to exploit the temporal information across frames and also leverage the predictions of the introspection function in motion planning to reduce the probability of failures for V-SLAM.